\documentclass[a4paper,fleqn]{cas-dc}
\usepackage[authoryear]{natbib}

\usepackage{hyperref}
\usepackage{booktabs}
\usepackage{multirow}
\usepackage{algorithm}
\usepackage[noEnd=false]{algpseudocodex}

\ExplSyntaxOn \cs_gset:Npn \__first_footerline: { \group_begin: \small \sffamily \__short_authors: \group_end: } \ExplSyntaxOff

\begin{document}
\let\printorcid\relax
\renewcommand{\topfraction}{0.99}
\renewcommand{\bottomfraction}{0.99}
\renewcommand{\textfraction}{0.01}
\renewcommand{\floatpagefraction}{0.99}

\shorttitle{Modern Neural Networks: The New Default for Digital Soil Mapping}
\shortauthors{Barkov et al.}

\title[mode = title]{Modern Neural Networks for Small Tabular Datasets: The New Default for Field-Scale Digital Soil Mapping?}
\author[1,2]{Viacheslav Barkov}
\author[1,2]{Jonas Schmidinger}
\author[2]{Robin Gebbers}
\author[1,3]{Martin Atzmueller}

\affiliation[1]{organization={Osnabrück University, Joint Lab Artificial Intelligence and Data Science}, city={Osnabrück}, country={Germany}}
\affiliation[2]{organization={Leibniz Institute for Agricultural Engineering and Bioeconomy (ATB), Department of Agromechatronics}, city={Potsdam}, country={Germany}}
\affiliation[3]{organization={German Research Center for Artificial Intelligence (DFKI), Research Department Cooperative and Autonomous Systems (CAS)}, city={Osnabrück}, country={Germany}}

\begin{abstract}
  In the field of pedometrics, tabular machine learning is the predominant method for soil property prediction from remote and proximal soil sensing data, forming a central component of Digital Soil Mapping (DSM). At the field-scale, this pedometric modeling task is typically constrained by small training sample sizes and high feature-to-sample ratios, particularly when using spectroscopic data. Traditionally, these conditions have proven challenging for conventional deep learning methods. Classical machine learning algorithms, particularly tree-based models like Random Forest and linear models such as Partial Least Squares Regression, have long been the default choice for pedometric modeling within DSM. Recent advances in artificial neural networks (ANN) for tabular data challenge this view, yet their suitability for field-scale DSM has not been proven. We introduce a comprehensive benchmark that evaluates state-of-the-art ANN architectures, including the latest multilayer perceptron (MLP)-based models (TabM, RealMLP), attention-based transformer variants (FT-Transformer, ExcelFormer, T2G-Former, AMFormer), retrieval-augmented approaches (TabR, ModernNCA), and an in-context learning foundation model (TabPFN). Our evaluation encompasses 31 field- and farm-scale datasets containing 30--460 soil samples and three critical soil properties: soil organic matter or soil organic carbon, pH, and clay content. Our results reveal that modern ANNs consistently outperform classical methods on the majority of tasks, demonstrating that deep learning has matured sufficiently to overcome the long-standing dominance of classical machine learning for soil property prediction. Notably, TabPFN delivers the strongest overall performance, showing robustness across varying conditions. We therefore recommend the adoption of modern ANNs for field-scale DSM and propose TabPFN as the new default choice in the toolkit of every pedometrician.
\end{abstract}


\nonumnote{Notice: The published version of this article is available in \textit{European Journal of Soil Science} at: \href{https://doi.org/10.1111/ejss.70299}{https://doi.org/10.1111/ejss.70299}}

\maketitle

\section{Introduction}
\label{section:introduction}

Soil maps are relevant for a range of environmental and agricultural applications~\citep{keesstra2016forum}. Especially at the field-scale, they play a crucial role in supporting high crop yields while considering negative environmental impacts~\citep{gebbers2010precision, rossel2016soil}. Sensors enable rapid and cost-effective generation of high-resolution soil data needed for creating these soil maps. This includes off-site sensing in the laboratory or direct in-situ measurements in field-conditions through proximal-soil sensing (PSS) or remote sensing (RS)~\citep{rossel2016soil}. However, these sensors do not directly measure actual soil properties relevant for agriculture but proxies related to them~\citep{gebbers2018proximal}. The translation of sensor signals into useful soil property estimates requires a set of soil samples to train a site-specific prediction model. This pedometric modeling step is central to Digital Soil Mapping (DSM), the creation of spatially referenced soil information through quantitative models that infer the spatial variations of soil properties from soil observations and high-resolution environmental covariates~\citep{mcbratney2003digital,mcbratney2019pedometrics}.

Tabular machine learning (ML) has emerged as the predominant approach for pedometric modeling in contemporary DSM ~\citep{heuvelink2022spatial}. However, the application of ML for field-scale DSM faces specific challenges, distinguishing it from other ML domains. These challenges include high noise, complexity of soil as a medium, variable measurement conditions, differences in measurement footprints and spatial autocorrelation. Most importantly, training sample sizes are generally rather low as soil sampling and soil analyses in laboratory rapidly becomes cost- and labor-prohibitive ~\citep{soderstrom2016adaptation}. This creates a fundamental tension: while datasets cannot be too small as this would result in poorly fitted prediction models, economic considerations constrain the size of training sample sets~\citep{schmidinger2024effect}. Additionally, the feature dimensionality in DSM can vary drastically from very low dimensional data generated by common in-situ PSSs like electrical conductivity sensors to very high dimensional data obtained from mid- and near-infrared spectroscopy. In soil spectroscopy, the feature dimensionality usually exceeds the number of soil samples, making modeling prone to overfitting~\citep{dos2023improving,wang2022determination,wang2022spectral}, a challenge known as the curse of dimensionality~\citep{bellman1957dynamic}.

These dataset characteristics favored classical ML approaches like tree-based algorithms and linear models, which became the default choices for pedometric modeling within DSM~\citep{wadoux2020machine,barra2021soil,ding2025advancing}. In particular, Random Forest dominates due to its straightforward application, good performance with default hyperparameters~\citep{probst2019hyperparameters}, fast training times, and general effectiveness even with low training sample sizes~\citep{ma2020comparison,schmidinger2024effect}. Recent reviews consistently report Random Forest as the most frequently used algorithm for soil property prediction in DSM~\citep{ding2025advancing,wadoux2020machine}. In soil spectroscopy applications specifically, linear models, usually combined with linear feature transformations such as principal component analysis (PCA) or partial least squares, have established themselves as fundamental methods~\citep{barra2021soil}. They demonstrate robust performance in contemporary spectroscopy studies, even when evaluated alongside popular ML algorithms such as Random Forest or Artificial Neural Networks (ANN)~\citep[e.g.,][]{xue2023validity,schmidinger2025limesoda}.

Deep learning applications in the form of various ANN architectures, notably Multilayer Perceptrons (MLP) and Convolutional Neural Networks, have attracted increasing interest for DSM~\citep{wadoux2025artificial}, but have mostly shown success for contexts where abundant training data is available. This includes large-area soil mapping at a regional up to continental level utilizing RS~\citep[e.g.,][]{yang2021regional} or vast spectral libraries~\citep[e.g.,][]{ng2019convolutional}. However, this success did not extend to field- or smaller regional-scale applications where sample sizes are commonly lower, with ANNs showing relatively poor performances~\citep[e.g.,][]{xue2023validity,oukhattar2025variability}. Moreover, even in large-area applications, deep learning did not always show improvements over classical methods~\citep[e.g.,][]{sun2023digital}. These limitations, particularly in smaller area high resolution soil mapping, are further emphasized by the methodological comparison of~\cite{khaledian2020selecting}, who argue that classical ML methods provide more stable results than ANNs in cases of limited training data. In fact,~\cite{khaledian2020selecting} explicitly recommend using classical ML algorithms when working with fewer than 1,000 soil samples. Overall, due to the uncertainty surrounding the effectiveness of deep learning for pedometric modeling in DSM, ANNs have not replaced classical ML approaches.

The historical dominance of classical ML in DSM aligns with broader trends in the tabular data domain, where classical ML approaches have maintained a competitive edge over ANNs until recent years. Following deep learning's widespread success in computer vision and natural language processing, early attempts to apply ANNs to tabular data generated considerable optimism, with several studies reporting promising results~\citep{arik2021tabnet, katzir2020net}. However, subsequent benchmarking studies revealed that classical ML, particularly tree-based ensemble methods like gradient-boosted decision trees (GBDTs), consistently outperformed these early deep learning approaches~\citep{shwartz2022tabular}. In fact, \citet{shwartz2022tabular} observed that the early claims of ANN superiority were largely driven by limited benchmarking practices, with studies relying on favorable results from a narrow selection of datasets, failing to generalize across broader evaluations. This was subsequently confirmed by other comprehensive tabular benchmarks~\citep{grinsztajn2022tree, mcelfresh2023neural}. Notably, these benchmarks confirmed classical ML superiority in conditions with limited sample sizes and high dimensionality. \citet{mcelfresh2023neural} observed that the performance gap between tree-based models and ANNs widened considerably as the dataset size decreased, specifically noting that ANNs performed comparatively worse when the feature-to-sample ratio became larger.

Recent developments in deep learning for tabular data challenge this perspective. Over the past two years, multiple ANN approaches claiming to surpass classical ML methods have been proposed for tabular data~\citep{hollmann2025accurate,holzmuller2025better,gorishniy2025tabm,ye2025revisiting,cheng2024arithmetic}. In contrast to previous deep learning models, recent tabular benchmarks also support this shift~\citep{erickson2025tabarena,ye2025closerlooktabpfnv2}. These advances extend far beyond incremental improvements to the classically used ANN architectures like MLPs. Entirely new tabular ANN approaches have emerged, including attention-based models, retrieval-based approaches, and in-context learning foundation models~\citep{ye2024closer}. We refer to these recent architectural innovations, from the past year, collectively as modern ANN approaches throughout this manuscript. These modern ANNs hold particular promise for addressing pedometric modeling challenges in DSM. In-context learning models such as TabPFN, a Tabular Prior-data Fitted Network~\citep{hollmann2025accurate}, are trained once on synthetic data to make predictions on new datasets from a context set of labeled examples provided in the input. They claim to achieve particularly strong performance on small datasets, making them potentially strong choices for the small-area DSM applications. Retrieval architectures such as ModernNCA~\citep{ye2025revisiting} exploit sample neighborhood relationships, and may offer unique advantages for soil property prediction given the spatially correlated nature of soil data.

Preliminary ML benchmarking studies showed impressive performance on public domain-independent benchmarks with these modern ANN approaches ~\citep{ye2024closer,erickson2025tabarena,hollmann2025accurate,ye2025closerlooktabpfnv2}. However, these studies did not adequately represent the specific constraints of field-scale DSM, as their dataset properties differed substantially from those encountered in field-scale DSM (see Table~\ref{table:benchmark-studies-review} in the Supporting Information~\ref{appendix:benchmark-studies-review}). For example, datasets in the benchmark of~\cite{erickson2025tabarena} and~\cite{ye2024closer} only included datasets with at least 500 observations. While this sample size is considered small in the broader ML context, it still largely exceeds the maximum sample size commonly available in field-scale DSM~\citep{schmidinger2025limesoda}. Additionally, the aspect of high-dimensionality and unfavorable feature-to-sample ratios is often dismissed or marginally discussed. Consequently, there remains a crucial need to examine the applicability and efficacy of these advanced deep learning paradigms in the context of the inherent constraints of field-scale DSM.

Applying and comparing these promising architectures in field-scale DSM is non-trivial. As the broader tabular data literature already warns~\citep{shwartz2022tabular}, proper evaluation of the new deep learning methods is itself a significant methodological challenge, and inadequate benchmarking practices can substantially affect conclusions about their relative performance. Methodological problems of previous tabular benchmarking studies have been largely overlooked in pedometrics, leading to critical limitations that may produce incomplete or misleading results~\citep{schmidinger2025limesoda}. Recent work by \citet{schmidinger2025limesoda} revealed that over 95\% of DSM benchmarking studies relied on a single dataset, with the largest number of datasets used in any reviewed study being only three. As demonstrated by~\cite{shwartz2022tabular}, reliance on a single dataset may inadvertently result in overinterpretation of incidental findings, obscuring the actual strengths and limitations of deep learning methods for pedometric modeling within DSM. Other confounding factors, like the choice of hyperparameters, can also introduce significant bias~\citep{niessl2022over}. In some DSM studies, hyperparameters of ANNs were carefully selected, whereas the competing tree-based models have not been tuned or no information regarding their optimization was provided~\citep[e.g.,][]{ng2019convolutional,yang2021regional}. Given that fewer than 10\% of DSM benchmarking studies provided open datasets and less than 5\% shared their code~\citep{schmidinger2025limesoda}, it is usually not possible to reproduce and verify the results. These shortcomings highlight the critical need for comprehensive benchmarking studies in DSM that systematically assess novel soil property prediction methods across multiple diverse datasets in a fair and rigorous way. Different training regimes, parameter optimization approaches, validation strategies, and other methodological choices must be implemented consistently to ensure fair comparisons~\citep{niessl2022over}.

In this work, we benchmark modern deep learning methods that have shown promise in tabular data applications but remain unexplored in the domain of pedometrics and DSM. Our study makes several key contributions. We are the first to systematically introduce and assess state-of-the-art deep learning methods for DSM at the field-scale with its distinct dataset challenges regarding sample size and feature dimensionality. Second, we establish a fair evaluation framework with consistent hyperparameter optimization, training regimes, and validation strategies across all methods. Third, we conduct experiments across multiple diverse field-scale soil datasets, avoiding the single-dataset evaluation dominating in previous benchmarking studies in pedometrics. Finally, we ensure complete reproducibility by making all datasets, code, and experimental configurations publicly available (see \hyperref[section:code-data-availability]{Code and data availability}). Through this analysis, we aim to provide empirical evidence on whether these advanced deep learning methods can overcome the longstanding dominance of classical ML in pedometric applications, particularly under the small-sample constraints typical for farm-scale DSM.

\section{Materials and methods}
\label{section:materials-and-methods}

\subsection{Datasets}
\label{section:datasets}

\begin{figure*}[htbp]
  \centering
  \includegraphics[width=1.0\textwidth]{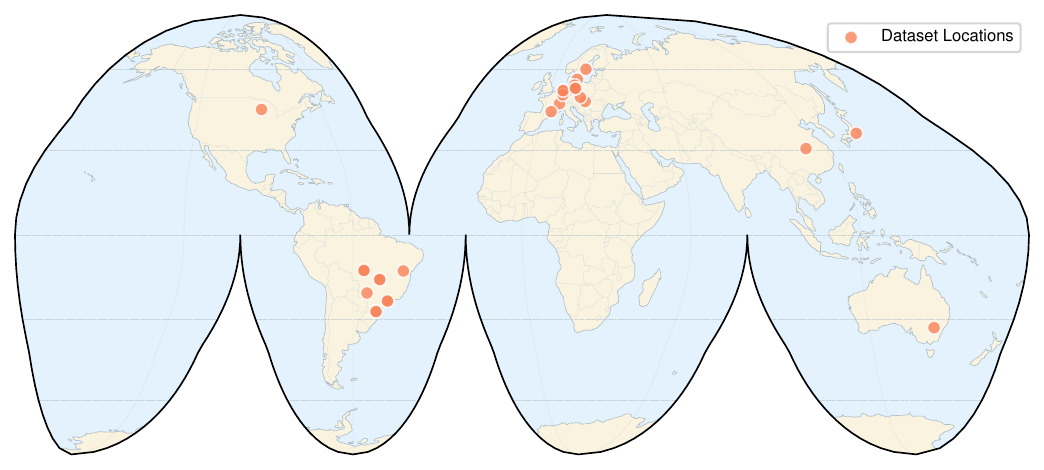}
  \caption{Global distribution of the 31 field- and farm-scale datasets used in this study. Background map is based on Homolosine projection. The datasets span multiple continents and diverse agricultural contexts, including locations across North and South America, Europe, Asia, and Australia. Base map source: Natural Earth.}
  \label{figure:datasets-map}
\end{figure*}

We utilized Precision Liming Soil Datasets (LimeSoDa), a collection of 31 field- to farm-scale datasets spanning multiple countries and diverse agricultural contexts, representing a broad spectrum of soil mapping scenarios encountered in precision agriculture~\citep{schmidinger2025limesoda}. Their global distribution is illustrated in Figure~\ref{figure:datasets-map}. Each dataset contains three target soil properties: soil organic matter (SOM) or soil organic carbon (SOC), pH, and clay content, yielding a total of 93 regression tasks for model evaluation. Individual dataset sizes range from 30 to 460 soil samples, providing varied data scenarios typical of real-world DSM projects for small-area soil mapping. All datasets are openly accessible and have been standardized in tabular format to facilitate reproducible research through the LimeSoDa repository (see \hyperref[section:code-data-availability]{Code and data availability}).

Features are dataset-specific and were obtained by different sensing technologies, including laboratory-based spectroscopy, in-situ PSS, and RS. This includes a variety of common sensor modalities, such as apparent electrical resistivity, gamma-ray spectrometry, ion selective electrodes, digital elevation models, multispectral RS data, vegetation indices, and X-ray fluorescence derived elemental concentrations. Several datasets incorporate high-dimensional optical spectroscopy data from visible and near-infrared (vis-NIR), near-infrared (NIR), and mid-infrared (MIR) spectrometers. These measurements result in feature sets with up to 2,489 variables, as the reflectance is typically recorded across numerous continuous wavelength bands. There are no categorical features present across any of the datasets. See Table~\ref{table:benchmark-studies-review} in the Supporting Information~\ref{appendix:benchmark-studies-review} for comparisons to feature sizes and feature-to-sample ratios reported in prior benchmarks.

\begin{figure}[t]
  \centering
  \includegraphics[width=0.45\textwidth]{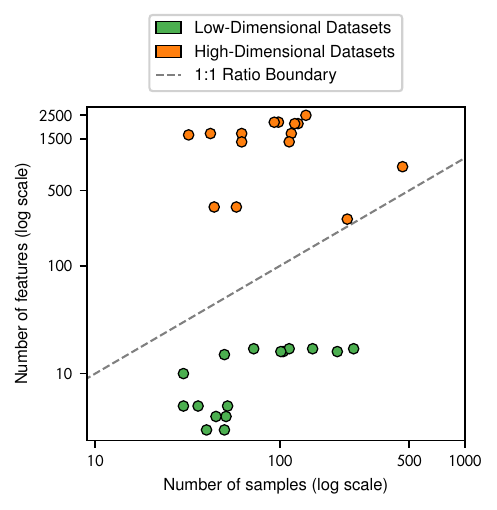}
  \caption{Distribution of datasets by number of features and soil samples (log scale), highlighting feature-to-sample ratios. Each point represents one dataset, with two groups shown: Low-Dimensional datasets (lower feature counts) and High-Dimensional datasets (higher feature counts due to spectral features).}
  \label{figure:datasets-size}
\end{figure}

Given the computational challenges posed by high-dimensional spectral data, we categorized the datasets into two groups based on feature characteristics and dimensionality for our analysis (Figure \ref{figure:datasets-size}). The first group, referred to as "High-Dimensional", includes datasets with feature-to-sample ratios $>$1, containing vis-NIR, NIR, or MIR spectroscopy features. These datasets require dimensionality reduction because the unfavorable feature-to-sample ratio would otherwise lead to severe overfitting~\citep{schmidinger2025limesoda}. Dimensionality reduction was performed during preprocessing as described in Section~\ref{section:data-preprocessing}. The second group, referred to as "Low-Dimensional", includes datasets with lower-dimensional PSS features (e.g., apparent electrical resistivity) or RS features (e.g., vegetation indices). All of these datasets have a feature-to-sample ratio $<$1.

\subsection{Classical machine learning models}
\label{section:classical-ml-baselines}

We included several algorithms that have demonstrated strong performance in pedometric modeling within DSM as classical ML baselines. Specifically, we evaluated Linear Regression (including regularized variants), Partial Least Squares Regression (PLSR), Random Forest, and XGBoost. Implementation details are provided in the Supporting Information~\ref{appendix:implementation-details}.

Linear Regression is a statistical method that models the relationship between features and target variables using a linear equation. When multiple independent features are used, it is defined as Multiple Linear Regression (MLR). Linear Regression, specifically MLR, serves as a fundamental baseline in DSM soil property prediction despite its simplicity, often serving as a primary baseline for comparison against other ML algorithms~\citep{wadoux2025artificial, ding2025advancing, wadoux2020machine}. In addition to MLR, we included its regularized variants in our evaluation. Specifically, we included Lasso regression (MLR with L1 norm regularization) and Ridge regression (MLR with L2 norm regularization).

PLSR is a statistical method that projects both features and target variables into a lower-dimensional latent space that maximizes their covariance and then fits a linear regression, aiming to enable modeling under severe multicollinearity and very large feature-to-sample ratios. We employed PLSR for High-Dimensional group of datasets, which requires dimensionality reduction prior to model fitting. For all other models we applied PCA as a preprocessing step for these datasets (see Section~\ref{section:data-preprocessing}).

Random Forest \citep{breiman2001random} is an ensemble of decision trees (DT) that uses bootstrap aggregating (bagging) to reduce the variance of a single DT by employing a randomized ensemble. It is the most widely adopted ML algorithm in DSM soil property prediction~\citep{ding2025advancing,wadoux2020machine}. \citet{wadoux2020machine} found that 80\% of reviewed studies employed at least one tree-based algorithm, with Random Forest being the most popular.

XGBoost~\citep{chen2016xgboost} is a popular implementation of GBDTs that has exhibited strong performance in many tabular data domains, including DSM. Unlike Random Forest, which trains DTs in parallel, GBDTs work by sequentially training DTs, with each new tree correcting the residuals of the previous ensemble. The optimization is performed by minimizing a differentiable loss function using gradient descent. GBDTs, and XGBoost in particular, have historically outperformed classical ML and early neural network approaches in general tabular regression benchmarks~\citep{grinsztajn2022tree}, although seeing more limited application in DSM compared to the widespread adoption of Random Forest~\citep{wadoux2025artificial}.

\subsection{Artificial neural networks}
\label{section:ann-architectures}

The landscape of deep learning for tabular data has evolved substantially in recent years, with various architectural innovations attempting to bridge the performance gap with traditional tree-based methods \citep{hollmann2025accurate,holzmuller2025better,gorishniy2025tabm,gorishniy2023revisiting,ye2025revisiting}. We included four groups of modern ANN architectures for tabular data, selecting established baseline models from each group alongside recent improvements introduced within the past year. Implementation details, model dependencies, and a concise glossary of ANN terminology are provided in the Supporting Information~\ref{appendix:implementation-details} and \ref{appendix:glossary}.

\subsubsection{MLP-based models}
\label{section:ann-mlp-based}

MLP represents a classical ANN baseline for tabular data. MLPs are not new to the domain of DSM and have been used and evaluated in previous studies, although generally underperforming compared to classical ML methods in field-scale DSM. We included a standard MLP as a baseline along with two recent architectural improvements, RealMLP and TabM.

MLP architecture consists of fully connected layers with non-linear activation functions. As noted by \citet{gorishniy2025tabm} and further elaborated by \citet{holzmuller2025better}, MLPs remain competitive baselines for tabular data, often matching or outperforming more complex attention- and retrieval-based architectures when evaluated under consistent experimental protocols. According to \citet{gorishniy2025tabm}, "MLPs [\dots] form a line of stronger and more practical models compared to attention- and retrieval-based architectures".

TabM~\citep{gorishniy2025tabm} extended the MLP architecture with new scaling parameter-efficient layers. The idea is to produce multiple predictions per input instance by training specialized weight-sharing components within a single model, retaining most parameters in common while still maintaining diversity through distinct prediction pathways.

RealMLP~\citep{holzmuller2025better} took a different approach, with contributions not limited to model architecture. \citet{holzmuller2025better} introduced an improved MLP version employing neural tangent parametrization, parametric Mish activation functions, and specialized scaling layers, with an important contribution of their own training regime and pre- and post-processing pipeline. The whole range of these contributions, which are not limited to ML architectural decisions, may significantly improve MLPs in the context of tabular data.

\subsubsection{Retrieval-based models}
\label{section:ann-retrieval-based}

We included retrieval-based models that utilize learned embedding spaces to find relevant training examples for guiding predictions, drawing inspiration from classical nearest neighbor approaches. We evaluated TabR, which pioneered this concept, and ModernNCA, introduced last year as an advancement of the approach.

TabR~\citep{gorishniy2023tabr} implemented this concept by searching for K nearest neighbors in a learned embedding space, computing their contributions based on both feature and label representations. These contributions are then aggregated with attention weights determined by distances in embedding space, effectively creating a differentiable variant of k-nearest neighbors with adaptive representation learning.

ModernNCA~\citep{ye2025revisiting} improved on the retrieval-based approach of TabR by revisiting Neighborhood Components Analysis (NCA) in a modern context. ModernNCA learns a deep embedding with a soft nearest-neighbor objective and predicts by distance-weighted averaging of neighbor labels in the learned embedding space. Training is made scalable with stochastic neighborhood sampling that evaluates neighbors against a random subset per mini-batch and the full set at inference.

\subsubsection{Attention-based models}
\label{section:ann-attention-based}

Inspired by the success of the transformer architecture in natural language processing~\citep{vaswani2017attention}, a range of attention-based models has emerged for tabular data, seeking to explicitly model feature interactions via self-attention. We included AutoInt as an early baseline, FT-Transformer as a more principled adaptation of the transformer architecture, and recent incremental improvements including T2G-Former, AMFormer, and ExcelFormer.

AutoInt~\citep{song2019autoint} was among the first to successfully adopt multi-head self-attention in a tabular setting. Each feature is treated as a separate token, and multiple self-attention layers capture higher-order interactions. Residual connections preserve the original feature representations, while deep stacking of attention layers iteratively refines the learned relationships among features.

FT-Transformer~\citep{gorishniy2023revisiting} provided a more principled adaptation of the transformer architecture to tabular data. Introducing a Feature Tokenizer module, it encodes numerical and categorical inputs into embeddings suitable for transformer processing. Unlike the original transformer, FT-Transformer introduced feature-specific biases and embedding schemes for numerical data. The rest of the architecture remains close to the standard transformer, featuring multi-head self-attention and feed-forward blocks.

Successive works on attention-based models have introduced multiple incremental upgrades to FT-Transformer. T2G-Former~\citep{yan2023t2gformer} incorporated feature-relation graphs to guide attention, while AMFormer~\citep{cheng2024arithmetic} integrated additive and multiplicative operations into the attention blocks, accounting for arithmetic feature interactions that commonly appear in tabular data. Meanwhile, ExcelFormer~\citep{chen2024excelformer} presented a transformer-based architecture with semi-permeable attention mechanisms, gating modules, and also contributed an improved training protocol, introducing new tabular data augmentation techniques.

\subsubsection{In-context learning}
\label{section:ann-in-context-learning}

We evaluated TabPFN, a foundation model that leverages in-context learning. It fundamentally differs from traditional ANN approaches by eliminating the need for dataset-specific training. Instead, it relies on pre-training on synthetic data and treats the usual training data as a context set provided to the model at inference time.

TabPFN~\citep{hollmann2025accurate} introduced the concept of Prior-data Fitted Network (PFN) for tabular data, pre-training on a vast collection of synthetic datasets sampled from a carefully designed prior. At inference time, TabPFN processes an entire dataset, including both training and test samples, in a single forward pass, effectively performing in-context learning. This approach is claimed to be particularly effective for small to medium-sized datasets (up to 10,000 observations), where TabPFN could achieve state-of-the-art performance without any dataset-specific training. Since TabPFN was originally released as a proof-of-concept classifier without regression support~\citep{hollmann2023tabpfnclassic}, its initial application for pedometrics and DSM became possible only after \citet{barkov2024efficient} introduced a target discretization approach to adapt it to regression tasks. The most recent TabPFN version~\citep{hollmann2025accurate} introduced several improvements, including a native regression capability, but has not been employed or evaluated for pedometric modeling within DSM prior to the present study.

\subsection{Experimental design}
\label{section:experimental-design}

\subsubsection{Fair architectural comparison}
\label{section:fair-comparison}

Our experimental protocol aimed to provide a fair comparison between different models. Recent advances in tabular deep learning often combine architectural innovations with complementary methodological improvements. For example, ExcelFormer introduces custom tabular-specific data augmentations, while RealMLP~\citep{holzmuller2025better} presents a comprehensive "bag-of-tricks" that extends well beyond architectural decisions to include specialized preprocessing (e.g., robust scaling with smooth clipping), postprocessing (e.g., output clipping), specific training procedures and best-epoch selection strategies. While these contributions are valuable, they are mostly model-agnostic and can theoretically be applied to any approach. We focused on unifying the evaluation protocol by either applying these improvements universally to every model, or excluding them from evaluation. This follows established conventions of fair comparison in tabular deep learning benchmarks, e.g.~ \citet{gorishniy2025tabm} who excluded custom data augmentations when evaluating against ExcelFormer performance to maintain experimental consistency.

Specifically, we applied robust scaling and universal numerical embeddings~\citep{gorishniy2023revisiting} to all models whenever possible. We used robust scaling for data preprocessing. We excluded interpolation-based augmentations employed by ExcelFormer. We did not include output clipping implemented in the postprocessing of RealMLP, as we do not want to prevent the models from extrapolating beyond the training data target variable range. We also employed systematic early stopping with a separate validation set. We provide further details of these decisions in the following sections. These choices allow us to benefit from modern training insights while preserving a balanced playing field for our model comparisons.

\subsubsection{Numerical feature embeddings}
\label{section:numerical-embeddings}

DSM commonly employs continuous numerical features derived from proximal and remote sensors. Traditionally, deep learning models represent these features as raw scalars. However, \citet{gorishniy2022embeddings} demonstrated that first transforming each scalar into a trainable vector, which they refer to as a numerical feature embedding, can significantly improve the performance of a plain MLP and allows even simple ANNs to become competitive with GBDTs on tabular benchmarks. As a consequence, modern tabular ANNs such as RealMLP, TabM, TabR and ModernNCA all implement numerical embeddings (usually piecewise-linear embeddings or their derivatives) as a core element of the network architecture. To eliminate confounding factors, we adopted a single embedding scheme for all models in this study that employ numerical feature embeddings. We used a piecewise-linear embedding module with a data-dependent number of quantile bins, implemented according to the reference code of \citet{gorishniy2022embeddings}.

\subsubsection{Data preprocessing}
\label{section:data-preprocessing}

We centered and scaled each feature by subtracting the median and dividing by the interquartile range (IQR), following~\citet{holzmuller2025better}. This procedure is commonly referred to as robust scaling. Unlike the mean and standard deviation typically used for scaling, the median and IQR are less sensitive to outliers, yielding more stable scaling statistics.

Specifically, for each column, let \(x_1,\dots ,x_n\in\mathbb{R}\) be the observed values and let \(q_p\) denote the \(p\)-quantile of \((x_1,\dots ,x_n)\) for \(p\in[0,1]\).

For every entry \(i\) we set
\[
  \begin{aligned}
    \tilde{x}_i  &= s\bigl(x_i-q_{1/2}\bigr),\\
    s &=
    \begin{cases}
      \dfrac{1}{q_{3/4}-q_{1/4}}, & \text{if } q_{3/4}\neq q_{1/4},\\[6pt]
      1, & \text{otherwise}.
    \end{cases}
  \end{aligned}
\]

Thus each feature is first centred by subtracting the median \(q_{1/2}\) and then scaled by the reciprocal interquartile range (IQR) whenever the IQR is non-zero. If the IQR vanishes, the scale factor is set to one to do centering only, avoiding division by zero.

For datasets whose spectroscopic feature dimensionality exceeds the number of soil samples we additionally applied PCA. PCA combined with MLR has repeatedly shown competitive performance in soil spectroscopy~\citep{barra2021soil,schmidinger2025limesoda}. \citet{schmidinger2025limesoda} identified PCA as the most effective preprocessing strategy when compared to using raw spectra or correlation-based feature selection in soil spectrometry. We therefore transformed the High-Dimensional datasets with PCA before model fitting, while keeping the raw variables for Low-Dimensional datasets. For datasets containing both spectroscopic and other sensor-derived features, PCA was applied to the combined feature set after robust scaling, thereby preserving potential covariance structures across sensor modalities. The number of retained components was treated as a hyperparameter and was optimized jointly with the model-specific parameters during the search described in Section~\ref{section:training-and-hyperparameters}.

All scaling statistics and PCA transformations were computed from the training set within each inner and outer fold to prevent data leakage.

\subsubsection{Validation}
\label{section:validation}

We applied nested cross-validation with 5 outer folds for evaluation and 5 inner folds for hyperparameter selection, consistent with common ML practices in DSM~\citep{kasraei2021quantile}. The complete algorithm detailing this is provided in Algorithm~\ref{alg:nested-cv} in the Supporting Information~\ref{appendix:nested-cv-algorithm}.

Following previous works, \citep[e.g.,][]{ye2025revisiting,gorishniy2023tabr}, we used patience-based early stopping on the inner validation set, while keeping the inner fold test set strictly for scoring. Prior studies typically used a patience threshold of 16 consecutive epochs without validation set improvement. We extended the threshold to 40 epochs to account for the very small dataset nature in field-scale DSM, since each epoch iteration would include fewer training observations. Some architectures, such as RealMLP, train for a fixed number of epochs, typically 256, and then select the best epoch according to the validation score, which effectively mirrors a higher-patience early-stopping scheme. Following this, for final model training, we used 256 training epochs with the best epoch determined by outer validation set, while keeping the outer fold test set strictly for final evaluation. This allows us to keep the advantages of patience-based stopping serving as a run pruning strategy during hyperparameter search, and have a final model trained to escape potential early stopping patience selection bias.

For early stopping, we used Root Mean Squared Error (RMSE):
\[
  \text{RMSE} = \sqrt{\frac{1}{n}\sum_{i=1}^{n}(y_i - \hat{y}_i)^2}
\]
where \(y_i\) are the true values, \(\hat{y}_i\) are the predicted values, and \(n\) is the number of observations.

For performance evaluation and ranking, we used R\textsuperscript{2} (coefficient of determination):
\[
  R^2 = 1 - \frac{\sum_{i=1}^{n}(y_i - \hat{y}_i)^2}{\sum_{i=1}^{n}(y_i - \bar{y})^2}
\]
where \(\bar{y}\) is the mean of the true values.

We ranked models based on their R\textsuperscript{2} scores for each regression task, and then averaged these ranks across datasets to obtain the overall performance metrics.

\subsubsection{Training and hyperparameter optimization}
\label{section:training-and-hyperparameters}

For hyperparameter optimization, we employed the tree-structured Parzen Estimator (TPE) algorithm~\citep{bergstra2011algorithms} with 100 rounds of optimization and an initial 20-round random search warm-up. This approach follows established best practices in recent tabular deep learning research. For example, \citet{gorishniy2025tabm} used TPE with 50-100 iterations across different model configurations. Our choice of 100 TPE rounds aligns with their findings, as the authors showed that 50-100 iterations typically provide adequate convergence for tabular models. We specifically utilized the multivariate TPE implementation~\citep{ozaki2022multiobjective} which accounts for dependencies between hyperparameters and has been shown to outperform independent parameter optimization in complex neural architectures.

We optimized parameters of all models (including classical ML algorithms) except for TabPFN, which does not require hyperparameter tuning. For ANN architectures, we explored hyperparameters such as the number of layers, hidden dimensions, dropout rates, and additional parameters dedicated to attention or retrieval mechanisms. For tree-based methods (XGBoost, Random Forest), we tuned maximum depth, subsampling ratios, and other regularization parameters. The number of trees was set to 1,000 for Random Forest, and XGBoost employed early stopping. The detailed hyperparameter grids can be found in Tables~\ref{table:ann-hyperparameters} and \ref{table:classical-ml-hyperparameters} in the Supporting Information~\ref{appendix:hyperparameter-grids}.

For datasets containing high-dimensional spectral features (vis-NIR, NIR, or MIR), we applied PCA and searched over 2 to 32 principal components, incorporating the choice of the number of components directly into the hyperparameter search. This consistent dimensionality reduction step was applied to all models to mitigate overfitting when the feature-to-sample ratio was highly unfavorable. Specifically for PLSR, we searched for the ideal number of components using the same range as for PCA.

All ANNs were trained using AdamW as the optimizer with mean squared error (MSE) loss. All random seeds were fixed to ensure reproducibility. We trained with a constant learning rate, which was optimized during hyperparameter search.

\subsubsection{Deep ensembles}
\label{section:deep-ensembles}

Our evaluation framework included deep ensembles to aggregate predictions from multiple independent training runs of each ANN architecture. State-of-the-art ANN methods routinely report deep ensemble results rather than single ANN performance. For instance, FT-Transformer and TabM report results as ensembles by averaging predictions from models trained with 15 different random seeds. Similarly, ExcelFormer uses 5 runs with different seeds per dataset, while TabPFN enforces ensembling by default using 8 estimators trained with feature permutations. Beyond ensuring fair evaluation, deep ensembles have strong theoretical foundations, with recent work demonstrating that independently trained ANNs improve test accuracy when data exhibits multi-view structure~\citep{allenzhu2023towards} and that diversity among ensemble members is crucial for generalization~\citep{jeffares2023joint}.

Furthermore, in the DSM context with extremely small datasets, we propose deep ensembles as a methodological solution to address data scarcity. Early stopping with a separate validation set requires sacrificing another portion of already limited training data. In our setup, this meant withholding 20\% of the data from training set within both inner and outer folds (see Section~\ref{section:training-and-hyperparameters}). Deep ensembles enable training multiple models not only with different weight initialization and data shuffling, but also on different validation splits, effectively ensuring that no single subset of data is excluded from training the deep ensemble while maintaining proper validation procedures for each individual model.

For fair evaluation across all methods, we fixed the ensemble size to 16 members for all ANN models. We conducted ablation studies to further analyze the contribution of deep ensembles to overall performance of ANNs in a DSM context (see Figure~\ref{figure:results-ablations-ensemble} in the Supporting Information~\ref{appendix:deep-ensembles}).

\section{Results}
\label{section:results}

\begin{figure*}[h]
  \centering
  \includegraphics[width=1.0\textwidth]{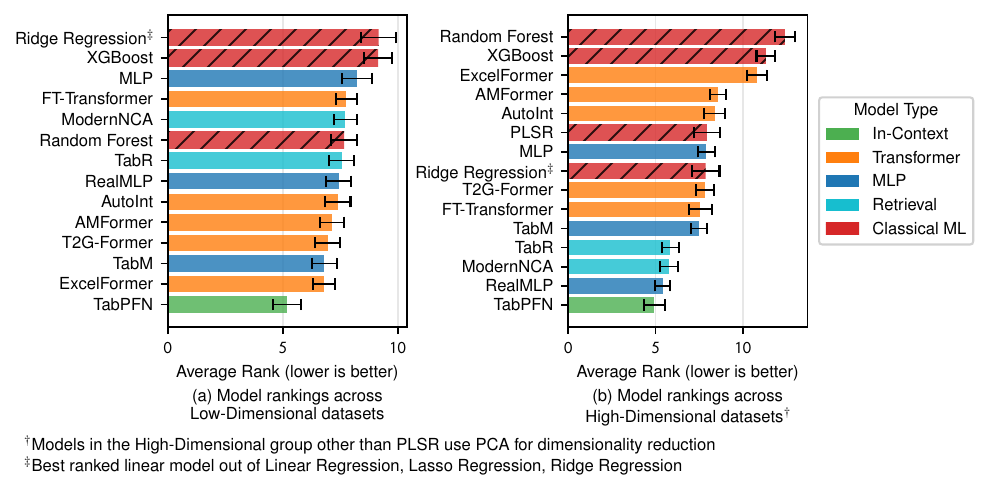}
  \caption{Performance comparison of ANN architectures versus classical ML baselines on DSM tasks. Models are grouped by architectural approach: MLP-based (MLP, TabM, RealMLP), Attention-based (AutoInt, T2G-Former, ExcelFormer), Retrieval-based (TabR, ModernNCA), and In-Context Learning (TabPFN). Lower rank values indicate better performance. Error bars show $\pm 1$ standard error of the mean of each model's rank across the datasets. The in-context learning model TabPFN consistently achieved the best performance across both Low-Dimensional and High-Dimensional datasets.}
  \label{figure:results-model-ranks}
\end{figure*}

The following results summarize model performance across 93 regression tasks spanning 31 datasets. The Low-Dimensional group comprised 16 datasets and 48 regression tasks, while the High-Dimensional group comprised 45 regression tasks from datasets with vis-NIR, NIR, and MIR spectroscopy features.

The performance of classical ML approaches and ANN architectures assessed through average R\textsuperscript{2} ranks is shown in Figure~\ref{figure:results-model-ranks}. Lower rank values indicate better performance, with the best possible rank being 1 and the worst being 14 for the Low-Dimensional group and 15 for the High-Dimensional group. Since we employed different regularization schemes for MLR (see Section~\ref{section:classical-ml-baselines}), the figure demonstrates the best-ranking linear model for each group. A detailed comparison of MLR, its regularized variants, and PLSR is provided in Figure~\ref{figure:linear-models-ranks} in the Supporting Information~\ref{appendix:linear-models-comparison}.

In the Low-Dimensional group, the in-context learning model TabPFN achieved the best overall performance. The strongest classical ML baseline was Random Forest. The attention-based ExcelFormer and MLP-based TabM emerged as the second-best performers. Other attention-based models, including T2G-Former, AMFormer, and AutoInt, also showed strong performance, outperforming Random Forest. However, three models failed to outperform Random Forest: MLP, FT-Transformer, and ModernNCA.

For High-Dimensional datasets, we additionally evaluated PLSR, while all other models in the High-Dimensional group used PCA as a preprocessing step (see Section~\ref{section:data-preprocessing}). The best-performing classical ML baseline for this group was MLR with Ridge penalty. While ANNs generally showed weaker absolute performance on these PCA-preprocessed datasets, most modern models still outperformed classical methods. TabPFN, similar to the Low-Dimensional group, showed the strongest performance. RealMLP ranked second after TabPFN, and retrieval-based models TabR and ModernNCA demonstrated notably stronger performance than on Low-Dimensional datasets. MLP and most attention-based models underperformed when compared to MLR with Ridge regularization. For the distributions of absolute $R^2$ scores across all tasks, refer to Figure~\ref{figure:appendix-r2-boxplots} in the Supporting Information~\ref{appendix:supplementary-results}.

\begin{figure*}[h]
  \centering
  \includegraphics[width=1.0\textwidth]{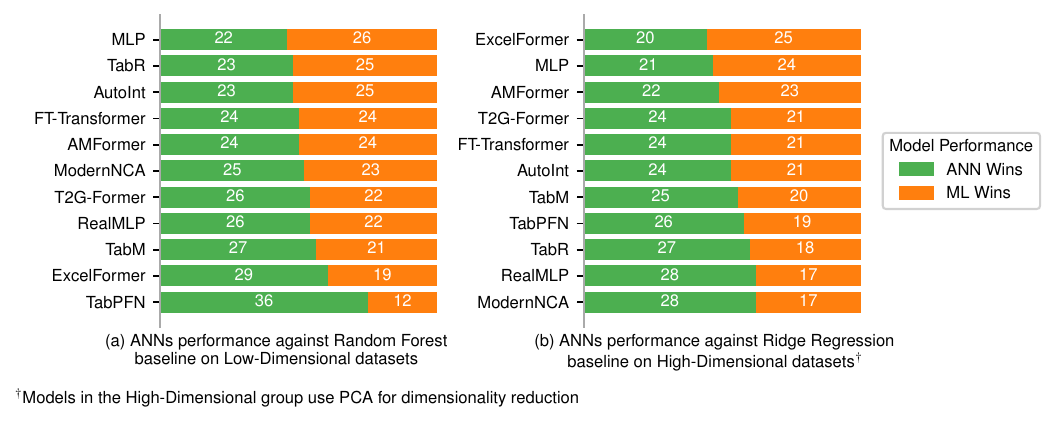}
  \caption{Head-to-head comparison of ANN architectures against the best-performing classical ML baseline for each dataset group. For Low-Dimensional datasets, models are compared against Random Forest; for High-Dimensional datasets, against Ridge Regression. A higher number of wins indicates superior performance. Modern deep learning approaches, particularly TabPFN, substantially outperformed classical baselines on the majority of tasks.}
  \label{figure:results-dl-vs-ml}
\end{figure*}

Figure~\ref{figure:results-dl-vs-ml} presents head-to-head comparisons between each ANN and the best classical baseline for its respective group. TabPFN achieved a 75\% win rate against Random Forest on Low-Dimensional tasks (36 wins, 12 losses). Other strong performers included ExcelFormer, TabM, RealMLP, and T2G-Former. On High-Dimensional tasks, three models achieved 62\% win rates against Ridge Regression: TabR, ModernNCA, and RealMLP. TabPFN followed closely with a 58\% win rate. Standard MLP consistently underperformed across both groups.

\begin{figure*}[h]
  \centering
  \includegraphics[width=1.0\textwidth]{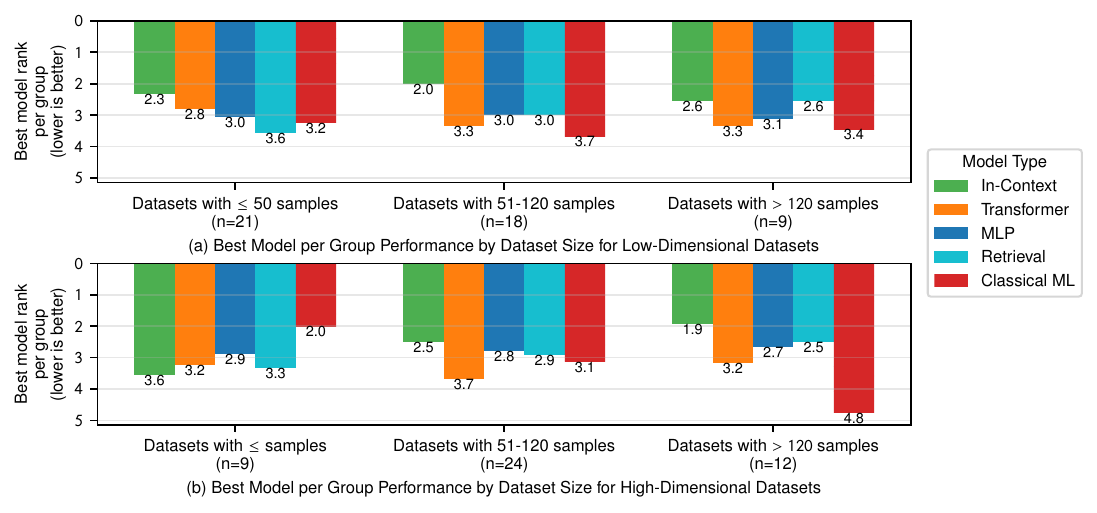}
  \caption{Model performance rankings grouped by dataset size, demonstrating the relationship between sample size and model effectiveness. Rankings are computed at the architectural group level using a two-stage process: first identifying the best model within each group, then ranking groups by their best model's performance. In-context learning (TabPFN) maintained superior performance across nearly all dataset sizes, with classical ML only prevailing on the smallest High-Dimensional datasets with PCA preprocessing.}
  \label{figure:results-model-ranks-by-dataset-size}
\end{figure*}

Figure~\ref{figure:results-model-ranks-by-dataset-size} reveals insights about the sample size-dependent performance. To examine this, we partitioned datasets into three categories based on sample size: $\leq$50 observations, 51--120 observations, and $>$120 observations. We employed two-stage ranking to avoid bias from varying numbers of models per model group. It shows that TabPFN maintained superior performance across nearly all scenarios, with classical ML surpassing it only for High-Dimensional datasets containing $\leq$50 observations. Retrieval-based models showed clear improvement with increasing dataset size in both groups, while MLP-based models exhibited stable performance regardless of size. Attention-based models showed degraded performance when PCA preprocessing was applied.

Overall, these results demonstrate that modern deep learning approaches, and TabPFN in particular, broadly surpass classical ML on the majority of field-scale DSM tasks.

\section{Discussion}
\label{section:discussion}

\subsection{Deep learning versus classical machine learning}
\label{section:discussion-dl-vs-ml}

Our results demonstrated, that modern ANNs outperformed classical ML methods on the majority of prediction tasks in LimeSoDa. While our results align with recent tabular benchmarking studies~\citep{ye2025closerlooktabpfnv2,erickson2025tabarena}, we extend their conclusions to the small-sample regime inherent to field-scale DSM, using datasets with as few as 30 soil samples. These small-sample conditions were historically perceived as unfavorable for ANNs~\citep{khaledian2020selecting,mcelfresh2023neural}, making our findings particularly notable for field-scale DSM.

The consistent performance of modern ANNs across both Low-Dimensional and High-Dimensional datasets directly challenges the established dominance of Random Forest and Linear Regression in pedometrics. Nonetheless, our results do not contradict previous findings that favored classical ML over ANN ~\citep{shwartz2022tabular,grinsztajn2022tree,mcelfresh2023neural,shmuel2024comprehensive} as standard MLP still underperformed compared to classical ML algorithms in field-scale DSM. The critical distinction between our findings and earlier ANN evaluations lies in modern ANN architectures that either enhance the MLP framework through architectural innovations (TabM, RealMLP) or introduce fundamentally different mechanisms such as attention (ExcelFormer, T2G-Former), retrieval (TabR, ModernNCA), or in-context learning (TabPFN).

Although modern ANNs outperformed classical ML on both Low-Dimensional and High-Dimensional datasets, the performance gap was notably smaller for High-Dimensional data. One key distinction in our High-Dimensional experiments was the use of PCA for dimensionality reduction, which was a straightforward solution to handle the unfavorable feature-to-sample ratios characteristic of spectroscopic data. While PCA has proven robust and superior to using full raw spectral data in LimeSoDa~\citep{schmidinger2025limesoda}, alternative dimensionality reduction strategies based on feature transformation (e.g., autoencoders~\citep{rumelhart1986autoencoders}), feature reduction (e.g., stochastic gates~\citep{yamada2020feature}) or ensemble-based feature subsampling~\citep{ye2025closerlooktabpfnv2} could further enhance ANN performance. However, comprehensive evaluation of such techniques for soil spectroscopy warrants dedicated investigation beyond the scope of this study. Regardless of these avenues for improvement, modern ANNs maintained their performance advantage and demonstrated general compatibility with dimensionality reduction techniques like PCA. This versatility across different data characteristics represents another key strength of selecting modern ANN architectures for pedometric modeling within DSM. It is a key advantage over tree-based models, which struggle with linearly transformed features such as those produced by PCA~\citep{menze2009comparison,grinsztajn2022tree,schmidinger2025limesoda}.

Our analysis extended the preliminary LimeSoDa benchmark of \citet{schmidinger2025limesoda}, which excluded ANNs due to their historically poor performance in tabular benchmarks. Their conclusion that different models excel in different scenarios remains partially valid. For example, we observed classical ML to be superior in the case of High-Dimensional datasets with extremely limited samples ($\leq$50 observations), as shown in Figure~\ref{figure:results-model-ranks-by-dataset-size}. However, beyond this exception, modern ANNs demonstrated remarkable consistency across various field-scale DSM scenarios. TabPFN in particular maintained superior performance across nearly all dataset sizes and dimensionalities.

The predictive performance improvements that modern ANNs demonstrate have immediate implications for pedometrics and DSM. DSM is fundamentally a task of model-based inference about unobserved locations, and its success depends on the ability of the pedometric model to represent the complex, often highly non-linear and cross-dependent, relationships between soil processes and the environment \citep{mcbratney2003digital}. With increasingly diverse PSS and RS covariates, modern ANNs appear more capable of capturing these dependencies. Moreover, these performance improvements matter for pedological understanding as well. Strong pedometric models, paired with model-agnostic interpretability methods \citep[e.g.,][]{lundberg2017shap}, have been suggested as a promising avenue for pedological knowledge discovery by generating hypotheses about soil-forming factors and mechanisms \citep{wadoux2020note}. In this context, high predictive accuracy is a prerequisite, as it increases the likelihood that the pedometric model has captured relevant underlying structures \citep{breiman2003statistical}, making modern ANNs a necessary step towards potential new insights in soil science.

\subsection{On the success of TabPFN}
\label{section:discussion-tabpfn}

TabPFN, a pioneer in tabular in-context learning, consistently achieved the highest performance across our experiments, emerging as the top-ranked model for both Low-Dimensional and High-Dimensional datasets. This aligns with \citet{hollmann2025accurate} claims that TabPFN specifically excels on small to medium-sized datasets. Nonetheless, previous tabular benchmarks demonstrated TabPFN's effectiveness mainly on datasets containing several hundred to thousands of observations (see Table~\ref{table:benchmark-studies-review} in the Supporting Information~\ref{appendix:benchmark-studies-review}).

The practical advantages of TabPFN extend beyond predictive performance. TabPFN is highly computationally efficient, producing predictions on evaluated DSM datasets in seconds. Although its transformer-based attention mechanism incurs a computational complexity of $O(n^2 + m^2)$ with respect to sample size $n$ and feature count $m$, this cost is negligible given the small datasets typical of field-scale DSM. Furthermore, its memory requirements scale only linearly as $O(n \cdot m)$~\citep{hollmann2025accurate}. Unlike conventional ML approaches, TabPFN requires neither hyperparameter optimization nor dataset-specific training. These characteristics eliminate two of the most time-consuming and technically challenging aspects of ML workflows~\citep{hollmann2025accurate}. This is especially significant in pedometric modeling, where the primary computational bottleneck often arises from iterative hyperparameter optimization rather than individual model training. Standard validation practices, such as nested cross-validation, further multiply the number of required training runs. TabPFN circumvents this bottleneck entirely, making it exceptionally practical for DSM applications. In fact, for pedometricians, this simplicity rivals that of Random Forest, which has long been valued for its reliable default performance and ease of application. The combination of superior accuracy and operational simplicity positions TabPFN as an ideal modeling approach for field-scale DSM. We therefore recommend TabPFN as a new default modeling choice for every pedometrician.

Additionally, TabPFN's success points toward opportunities for creating pedology-specific foundation models. The synthetic data prior underlying TabPFN encodes general causal structures that enable broad generalization across different tabular domains. However, soils exhibit distinctive properties with unique structure and complex biogeochemical interactions that generic synthetic priors may not fully represent. Exploring soil-informed priors that embed fundamental domain knowledge and physical rules represents a promising research direction \citep{minasny2024ssiml} and could result in specialized tabular foundation models for pedometrics. Furthermore, fine-tuning TabPFN with real soil datasets could produce models that better understand these distinctive patterns in pedological data \citep{rubachev2025finetuning,thomas2024retrieval}.

Despite these strengths, TabPFN's capabilities should not be overstated, as there are scenarios where its performance remains limited. These include datasets with very large training sets, high-dimensional feature spaces without dimensionality reduction, and multi-class classification tasks~\citep{hollmann2025accurate,rubachev2025tabred,ye2025closerlooktabpfnv2}. These constraints, however, fall outside the typical scope of field-scale DSM.

It is equally important to recognize that the advancements demonstrated in our study extend beyond TabPFN alone. Retrieval-based models like TabR and ModernNCA, MLP-based TabM and RealMLP, and attention-based models such as T2G-Former and ExcelFormer all consistently outperformed both classical ML baselines and classical ANN architectures like MLP. Each architecture brings unique strengths and design principles that could prove valuable for specific DSM applications or inform the design of future soil-specific models. We therefore interpret the success of TabPFN as a representative example of the broader capabilities that modern ANNs bring to DSM.

Finally, even the best-performing pedometric model remains subject to fundamental DSM constraints, and absolute predictive performance still varies substantially across the different LimeSoDa datasets (as shown in Figure~\ref{figure:appendix-r2-boxplots} in the Supporting Information~\ref{appendix:supplementary-results}). Advanced modeling techniques can reduce predictive uncertainty, but they represent only one component of the broader DSM workflow. Equally important are the representativeness of the sampling, influenced by the sampling design and sample size, and the quality of the features, affected by noise, preprocessing, and their overall relationship to the target soil property. Even the most advanced models will exhibit high uncertainty and poor extrapolation in under-sampled conditions, regardless of the choice of algorithm \citep{schmidinger2024effect,zizala2024soil,saurette2022effects}. Likewise, the set of features must adequately represent soil characteristics to allow the model to find meaningful patterns. We therefore regard TabPFN, and modern ANNs in general, as a substantial but not standalone advancement that complements, rather than replaces, the fundamental sampling and feature engineering steps of DSM.

\subsection{Future considerations}
\label{section:discussion-future-work}

The emergence of tabular in-context learning has catalyzed development of additional tabular foundation models, including MotherNet~\citep{mueller2024mothernet} and TabICL~\citep{jingang2025tabicl}. We did not include these models in our study, as they currently only support classification. However, they may be adapted to regression tasks employing the discretization approach proposed by \citet{barkov2024efficient}. Although this is beyond the scope of our current study, it represents a promising avenue for future research.

Beyond predictive accuracy, uncertainty quantification remains critical for pedometric modeling within DSM~\citep{breure2022loss,barkov2024efficient}. While our benchmark focused on point predictions, models like TabPFN can naturally provide probabilistic predictions for uncertainty estimation, since they perform regression using target discretization~\citep{hollmann2025accurate,barkov2024efficient}. Furthermore, other ANN-specific model-agnostic methods for uncertainty quantification can be evaluated, such as Laplace approximation~\citep{kristiadi2021laplace} or Monte Carlo dropout \citep{huang2025monte}.

Additionally, since we propose modern ANNs as a safe default choice for future pedometrical studies, this opens the field to ANN-specific improvements that could further boost model performance, such as model souping~\citep{wortsman2022modelsoups} or specialized regularization frameworks like TANGOS~\citep{jeffares2023tangos}.

Furthermore, the spatial nature of soil data presents compelling research directions beyond our tabular framing \citep{heuvelink2022spatial}. Future work could explore ways to incorporate spatial information into these high-performing tabular models, potentially through hybrid architectures that combine the demonstrated effectiveness of models like TabPFN with explicit spatial correlation structures.

These opportunities, alongside the proven performance of modern architectures, demonstrate that ANNs have matured sufficiently to not only advance field-scale DSM but also unlock the rapidly evolving deep learning methodological landscape for future advancements.

\section{Conclusion}
\label{section:conclusion}

This study provides the first systematic assessment of modern ANNs for small tabular datasets from DSM. We implemented and evaluated a comprehensive range of recent state-of-the-art ANN architectures, including MLP-based models (MLP, TabM, RealMLP), attention-based transformers (AutoInt, FT-Transformer, T2G-Former, ExcelFormer, AMFormer), retrieval-based approaches (TabR, ModernNCA), and in-context learning models (TabPFN). We compared these architectures against established classical ML baselines (Random Forest, XGBoost, Linear Regression) across 31 diverse datasets from LimeSoDa. Through this multi-dataset and fully reproducible benchmark, we establish a new comprehensive benchmarking standard for pedometrics and DSM.

Our results demonstrate that many modern ANNs consistently outperform classical ML methods in field-scale DSM tasks, even under challenging small-sample conditions. The in-context learning model TabPFN emerged as a particularly strong method, surpassing Random Forest, Linear Regression, and PLSR, the long-standing default methods in pedometric modeling within DSM. We therefore recommend TabPFN as the new default model and baseline for field-scale DSM. This recommendation is motivated not only by its superior predictive performance, but also by its operational simplicity, eliminating the need for hyperparameter optimization and dataset-specific training.

While classical ML retains advantages in specific areas, such as extremely small and high-dimensional datasets with fewer than 50 soil samples where PCA preprocessing is employed, its superiority diminishes rapidly as dataset size increases. Even for datasets containing only 50 to 100 soil samples, modern neural architectures can demonstrate clear performance advantages, challenging former views about sample size requirements for deep learning in general.

Our findings indicate that the recent wave of tabular deep learning research translates into tangible benefits for DSM. These results mark a substantial step forward for DSM, and as both deep learning and pedometrics fields continue to evolve, future advancements hold promise for further enhancing map accuracy and reliability in precision agriculture applications.

\section*{Code and data availability}
\phantomsection
\label{section:code-data-availability}

All the experiments are reproducible, and the source code, including the training and evaluation scripts, is open and publicly available at \url{https://github.com/slavabarkov/smalltabnets}. All the data used in this study is publicly available in the LimeSoDa repository~\citep{schmidinger2025limesoda} at \url{https://zenodo.org/records/14936177}.

\section*{Acknowledgements}
\phantomsection
\label{section:acknowledgements}

This work was supported by the Lower Saxony Ministry of Science and Culture (MWK), via the zukunft.niedersachsen program of the Volkswagen Foundation (ZN4072). Compute resources were funded by the Deutsche Forschungsgemeinschaft (DFG, German Research Foundation) project number 456666331.

\bibliographystyle{cas-model2-names}
\bibliography{references}

\onecolumn
\appendix

\makeatletter
\newcommand{\appsection}[1]{%
  \renewcommand{\@seccntformat}[1]{Supporting Information\ \csname the##1\endcsname:\ }
  \section{#1}
  \renewcommand{\@seccntformat}[1]{\csname the##1\endcsname}
}
\makeatother

\makeatletter
\newcommand{\appsubsection}[1]{%
  \renewcommand{\@seccntformat}[1]{Supporting Information\ \csname the##1\endcsname:\ }
  \subsection{#1}
  \renewcommand{\@seccntformat}[1]{\csname the##1\endcsname}
}
\makeatother

\setcounter{figure}{0}
\renewcommand{\thefigure}{S\arabic{figure}}
\setcounter{table}{0}
\renewcommand{\thetable}{S\arabic{table}}
\setcounter{algorithm}{0}
\renewcommand{\thealgorithm}{S\arabic{algorithm}}


\appsection{Detailed comparison of linear models}
\label{appendix:linear-models-comparison}

\begin{figure*}[htbp]
  \centering
  \includegraphics[width=\textwidth]{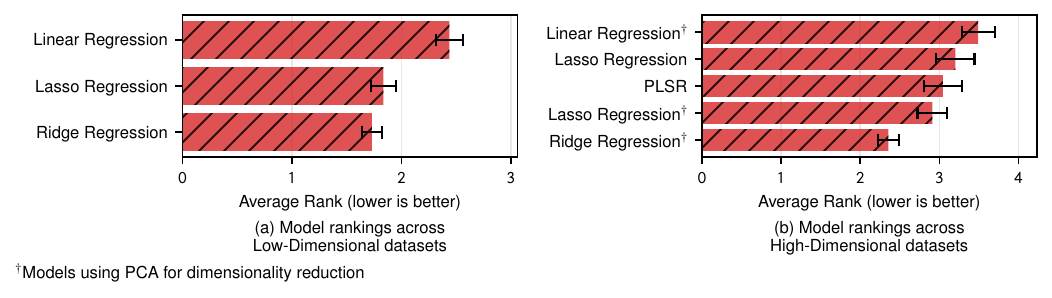}
  \caption{Average rank comparison of the four linear baselines across all 93 regression tasks.  Lower rank values indicate better performance (best = 1). Panel (a) shows results for Low-Dimensional datasets, panel (b) for High-Dimensional datasets that were reduced with PCA. Error bars denote $\pm$ one standard error of the mean.}
  \label{figure:linear-models-ranks}
\end{figure*}

The Figure~\ref{figure:linear-models-ranks} summarizes how the linear baselines rank across all 93 regression tasks used in this study.


\appsection{Deep ensembles ablation study}
\label{appendix:deep-ensembles}

\begin{figure*}[h]
  \centering
  \includegraphics[width=1.0\textwidth]{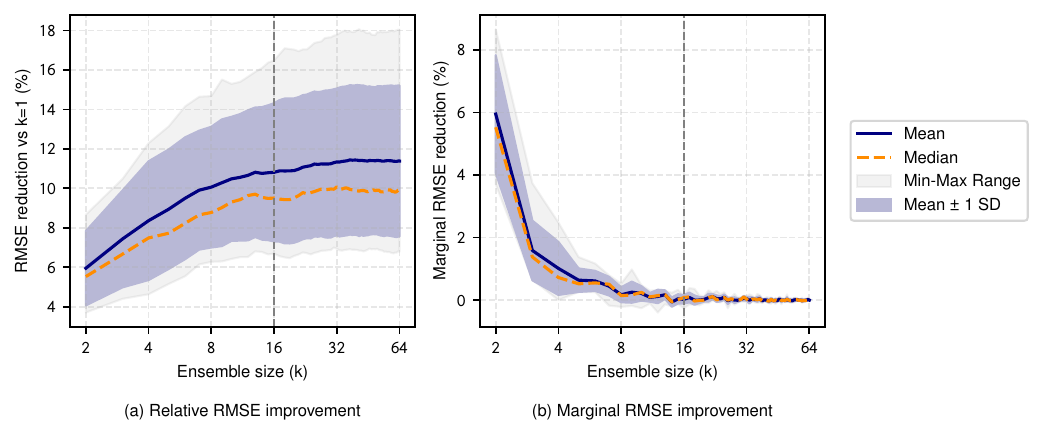}
  \caption{Impact of deep ensemble size on predictive performance, showing both absolute RMSE reduction and marginal improvements. Performance gains plateau after approximately 16 ensemble members, validating our choice of ensemble size and demonstrating that even small ensembles provide substantial benefits over single models in DSM.}
  \label{figure:results-ablations-ensemble}
\end{figure*}

To validate our choice of ensemble size, we conducted ablation studies varying the number of ensemble members from 1 to 64. Figure~\ref{figure:results-ablations-ensemble} presents these results.

Deep ensembles consistently improved predictive accuracy, but with diminishing returns beyond 16 members. Marginal gains decay rapidly after $k \approx 16$ and plateau after $k \approx 32$. This empirically supports the ensemble size used throughout the main experiments. Even small ensembles provide significant performance improvements over single models, highlighting the inherent instability of individual ANN training runs and demonstrating the value of deep ensembles for tabular data for a fair model evaluation.

\appsection{Nested cross-validation and model optimization procedure}
\label{appendix:nested-cv-algorithm}

Algorithm~\ref{alg:nested-cv} presents the complete nested cross-validation procedure employed in this study. The algorithm details the integration of hyperparameter optimization within inner folds, the prevention of data leakage through fold-specific scaling and dimensionality reduction, early stopping mechanisms, and the ensemble strategy for final predictions. The approach ensures that the outer test sets remain strictly independent from all model selection and training decisions.

\begin{algorithm}[h]
  \caption{Nested cross-validation with hyperparameter optimization and ensembling}
  \label{alg:nested-cv}
  \begin{algorithmic}
    \Require Dataset $D$, learning algorithm $A$, outer folds $K$, inner folds $N$, optimization trials $T$, ensemble size $M$
    \For{$i=1$ to $K$}
    \State Split $D$ into $D_{\text{train}}^i$ and $D_{\text{test}}^i$
    \For{$t=1$ to $T$}
    \State Sample hyperparameters $\theta_t$ from search space
    \For{$j=1$ to $N$}
    \State Split $D_{\text{train}}^i$ into $D_{\text{train}}^{i,j}$ and $D_{\text{val}}^{i,j}$
    \State Calculate scaling statistics on $D_{\text{train}}^{i,j}$, scale $D_{\text{train}}^{i,j}$ and $D_{\text{val}}^{i,j}$
    \If{dataset is high-dimensional (spectroscopy)}
    \State Fit PCA on $D_{\text{train}}^{i,j}$, transform $D_{\text{train}}^{i,j}$ and $D_{\text{val}}^{i,j}$
    \EndIf
    \State Train $A(\theta_t)$ on $D_{\text{train}}^{i,j}$ with early stopping on $D_{\text{val}}^{i,j}$, set $\text{RMSE}_{t,j}$ on $D_{\text{val}}^{i,j}$
    \EndFor
    \State Set $\text{RMSE}_t$ as average of $\text{RMSE}_{t,1}, \dots, \text{RMSE}_{t,N}$
    \EndFor
    \State Select best hyperparameters $\theta_i^\star$ with lowest $\text{RMSE}_t$
    \For{$m=1$ to $M$}
    \State Split $D_{\text{train}}^i$ into $D_{\text{train}}^{i,m}$ and $D_{\text{val}}^{i,m}$
    \State Calculate scaling statistics on $D_{\text{train}}^{i,m}$, scale $D_{\text{train}}^{i,m}$, $D_{\text{val}}^{i,m}$, and $D_{\text{test}}^{i}$
    \If{dataset is high-dimensional (spectroscopy)}
    \State Fit PCA on $D_{\text{train}}^{i,m}$, transform $D_{\text{train}}^{i,m}$, $D_{\text{val}}^{i,m}$, and $D_{\text{test}}^{i}$
    \EndIf
    \State Train $A(\theta_i^\star)$ on $D_{\text{train}}^{i,m}$ with early stopping on $D_{\text{val}}^{i,m}$
    \State Predict $D_{\text{test}}^i$ and set predictions $\hat{y}_{i,m}$
    \EndFor
    \State Set final predictions $\hat{y}_i$ as average of $\hat{y}_{i,1}, \dots, \hat{y}_{i,M}$
    \EndFor
    \State \Return Concatenated predictions $\hat{Y} = \{\hat{y}_1, \dots, \hat{y}_K\}$
  \end{algorithmic}
\end{algorithm}

\pagebreak

\appsection{Supplementary Results}
\label{appendix:supplementary-results}

\begin{figure*}[h]
  \centering
  \includegraphics[width=\textwidth]{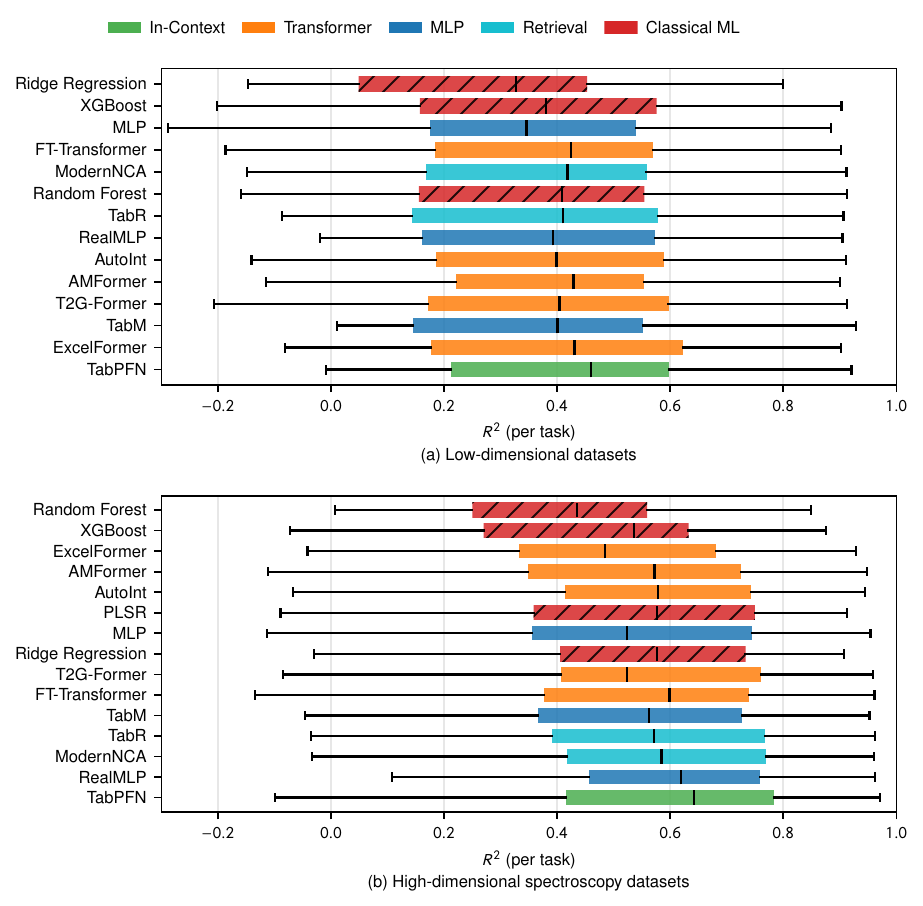}
  \caption{Distribution of absolute $R^2$ scores across all regression tasks, including results for (a) Low-Dimensional datasets and (b) High-Dimensional spectroscopy datasets. Models are ordered vertically by their average rank for cross comparison. Whiskers extend to $1.5\times$ the interquartile range.}
  \label{figure:appendix-r2-boxplots}
\end{figure*}

While rank-based aggregation provides a robust comparison of relative model performance across different regression tasks, it might abstract away the magnitude of differences across different models. Figure~\ref{figure:appendix-r2-boxplots} illustrates the distribution of absolute $R^2$ values for all 93 regression tasks across two dataset groups.

The distributions show that better ranking of modern ANNs is supported by concrete improvements in predictive accuracy. For Low-Dimensional datasets, TabPFN shows a notably higher median $R^2$ with a tighter interquartile range when compared against Random Forest. Similarly, in High-Dimensional spectroscopy tasks, TabPFN clearly outperforms the best ranking classical baseline, Ridge Regression, achieving both a higher median and higher maximum predictive performance. Retrieval-based models (TabR, ModernNCA) in High-Dimensional tasks show median performance comparable to Ridge Regression, but additionally exhibit noticeably higher upper-bound results.


\appsection{Benchmark studies review}
\label{appendix:benchmark-studies-review}

Table~\ref{table:benchmark-studies-review} summarizes the dataset properties used in prior influential tabular benchmarking studies. As shown, most benchmarks focused on datasets with significantly larger sample sizes and typically very low feature-to-sample ratios. Among these, only \citet{ye2025closerlooktabpfnv2} incorporated datasets that closely resemble our High-Dimensional setting as part of a broader benchmark. We did not include the dataset properties from \citet{mcelfresh2023neural}, because we only found publicly available information from a subset of datasets used in that benchmark.

\begin{table*}[h]
  \caption{Summary statistics of dataset properties from other previously mentioned benchmarks, compared to those used in our study, denoted as "ours". Most other benchmarks consist of datasets with larger sample sizes and low dimensional features.}
  \centering
  \label{table:benchmark-studies-review}
  \resizebox{\linewidth}{!}{%
    \begin{tabular}{lrrrrrrrrr}
      \toprule
      \textbf{Benchmark}
      & \multicolumn{3}{c}{\textbf{Sample}}
      & \multicolumn{3}{c}{\textbf{Feature}}
      & \multicolumn{3}{c}{\textbf{Feature-to-sample ratio}} \\
      \cmidrule(lr){2-4} \cmidrule(lr){5-7} \cmidrule(lr){8-10}
      & Min & Median & Max
      & Min & Median & Max
      & Min & Median & Max \\
      \midrule
      \citet{shwartz2022tabular}         & 7k   & 500k  & 1M    & 10    & 32    & 2k    & <0.001 & 0.001 & 0.009 \\
      \citet{grinsztajn2022tree}         & 2.6k & 17.4k & 940k  & 4     & 12    & 613   & <0.001 & 0.001 & 0.085 \\
      \citet{shmuel2024comprehensive}    & 43   & 4.8k  & 245k  & 4     & 13    & 267   & <0.001 & 0.004 & 0.116 \\
      \citet{ye2024closer}               & 506  & 5.1k  & 153k  & 3     & 16    & 308   & <0.001 & 0.003 & 0.161 \\
      \citet{hollmann2025accurate}       & 240  & 2.1k  & 10k   & 3     & 20    & 376   & 0.001  & 0.011 & 0.517 \\
      \citet{ye2025closerlooktabpfnv2}*  & 50   & 151   & 7k    & 2k    & 5.4k  & 22.3k & 0.714  & 41.807 & 262.153 \\
      \citet{ye2025closerlooktabpfnv2}\dag & 21k  & 626k  & 10M   & 8     & 54    & 784   & <0.001 & <0.001 & 0.025 \\
      \citet{erickson2025tabarena}       & 748  & 6.5k  & 150k  & 5     & 18    & 1.8k  & <0.001 & 0.005 & 0.474 \\
      \midrule
      High Dimensional (ours)            & 32   & 98    & 460   & 272   & 1.7k  & 2.5k  & 1.178  & 17.350 & 51.156 \\
      Low Dimensional (ours)             & 30   & 52    & 250   & 3     & 14    & 17    & 0.060  & 0.126 & 0.433 \\
      \bottomrule
    \end{tabular}
  }
  \footnotesize\flushleft
  *, \dag : \citet{ye2025closerlooktabpfnv2}, expands on \citet{ye2024closer} with an additional high-dimensional setting (*) and a large-sample setting (\dag).
\end{table*}


\appsection{Hyperparameter search space}
\label{appendix:hyperparameter-grids}

\appsubsection{Hyperparameters of artificial neural networks}
\label{appendix:hyperparameter-grids-ann}

\begin{table*}[h]
  \centering
  \small
  \caption{Search spaces for modern ANN architectures. Continuous intervals were sampled uniformly, logarithmic intervals log-uniformly, and width/dimension parameters were sampled from the categorical sets shown.}
  \label{table:ann-hyperparameters}
  \setlength{\tabcolsep}{4.5pt}
  \begin{tabular}{lccccccc}
    \toprule
    \multirow{2}{*}{Model}
    & Learning          & Weight         & Batch & \#\,Blocks / &
    Hidden            & Attention      & Dropout$^{\dagger}$ \\[-2pt]
    & rate\,(log)       & decay\,(log)   & size  & layers        &
    width (\(d\))     & heads          & range  \\
    \midrule
    AMFormer     & $[10^{-4},5\!\times\!10^{-3}]$ & $[10^{-6},10^{-2}]$ &
    \{4,8,16,32\} & 1--4 & 16--64 & 2,4 & 0--0.3 \\
    AutoInt      & $[10^{-4},5\!\times\!10^{-3}]$ & $[10^{-6},10^{-2}]$ & \{4,8,16,32\} & 1--4 & 16--64 & 2,4 & 0--0.3 \\
    ExcelFormer  & $[10^{-4},5\!\times\!10^{-3}]$ & $[10^{-6},10^{-2}]$ & \{4,8,16,32\} & 1--4 & 16--64 & 2,4 & 0--0.3 \\
    FT-Transformer & $[10^{-4},5\!\times\!10^{-3}]$ & $[10^{-6},10^{-2}]$ & \{4,8,16,32\} & 1--4 & 16--64 & 2,4 & 0--0.3 \\
    T2G-Former   & $[10^{-4},5\!\times\!10^{-3}]$ & $[10^{-6},10^{-2}]$ & \{4,8,16,32\} & 1--4 & 16--64 & 2,4 & 0--0.3 \\
    \midrule
    MLP          & $[10^{-4},5\!\times\!10^{-3}]$ & $[10^{-6},10^{-2}]$ & \{4,8,16,32\} & 2--4 & 64--512 & —   & 0--0.3 \\
    RealMLP      & $[10^{-4},5\!\times\!10^{-3}]$ & $[10^{-6},10^{-2}]$ & \{4,8,16,32\} & 2--4 & 64--512 & —   & 0--0.3 \\
    TabM         & $[10^{-4},5\!\times\!10^{-3}]$ & $[10^{-6},10^{-2}]$ & \{4,8,16,32\} & 2--4 & 64--512 & —   & 0--0.3 \\
    \midrule
    ModernNCA    & $[10^{-4},5\!\times\!10^{-3}]$ & $[10^{-6},10^{-2}]$ & \{4,8,16,32\} & 2--4 & 64--512 & —   & 0--0.3 \\
    TabR         & $[10^{-4},5\!\times\!10^{-3}]$ & $[10^{-6},10^{-2}]$ & \{4,8,16,32\} & Enc.\ 0--4, Pred.\ 1--4 &
    64--512 & — & 0--0.3 \\
    \bottomrule
  \end{tabular}
  \footnotesize\flushleft
  \dag : All dropout-type hyperparameters (e.g.\ attn\_dropout, ffn\_dropout, residual\_dropout, context\_dropout).
\end{table*}

Table~\ref{table:ann-hyperparameters} summarizes the hyperparameter search spaces of all neural architectures considered in this study, with the exception of TabPFN, which is a foundation model that requires no hyperparameter optimization. Parameters that are shared across similar models are listed under a common name. During hyperparameter optimization, all continuous ranges were sampled uniformly, all the ranges indicated with sub-script "log" were sampled log-uniformly, and all width and dimension parameters were drawn from the categorical sets shown in the table. For the High-Dimensional data group we applied PCA in a joint search, sampling the number of components uniformly from 2--32.

Some model-specific hyperparameters that are not listed in Table~\ref{table:ann-hyperparameters} were additionally optimized. For AMFormer these were number of prompt tokens $\in \{1,2,4\}$, additive and multiplicative attention tokens $\in \{1,2,4\}$. For ModernNCA these were retrieval parameters, including embedding dimensionality $64\text{--}256$ and soft-NN temperature $0.1\text{--}5.0$ with log-uniform sampling. For TabR these were retrieval parameters, including context size $\{2,4,8,16\}$ and context dropout $0\text{--}0.3$.

\appsubsection{Hyperparameters of classical machine learning algorithms}
\label{appendix:hyperparameter-grids-classical-ml}

\begin{table*}[h]
  \centering
  \caption{Hyperparameter grids for classical baselines. Bracketed intervals were sampled uniformly; "log" indicates log-uniform sampling.}
  \label{table:classical-ml-hyperparameters}
  \begin{tabular}{lll}
    \toprule
    \textbf{Algorithm} & \textbf{Parameter} & \textbf{Search grid} \\
    \midrule
    XGBoost & learning\_rate & $[0.001,\,0.3]_{\log}$ \\
    & max\_depth & $\{3,\dots,10\}$ \\
    & min\_child\_weight & $\{1,\dots,10\}$ \\
    & subsample & $[0.5,\,1.0]$ \\
    & colsample\_bytree & $[0.5,\,1.0]$ \\
    & gamma & $[0,\,5]$ \\
    & reg\_alpha & $[10^{-6},\,10]_{\log}$ \\
    & reg\_lambda & $[0.1,\,10]_{\log}$ \\
    \midrule
    Random\,Forest & max\_depth & $\{3,\dots,30\}$ \\
    & min\_samples\_split & $\{2,\dots,10\}$ \\
    & min\_samples\_leaf & $\{1,\dots,10\}$ \\
    & max\_features & $[0.6,\,1.0]$ \\
    \midrule
    Lasso\,Regression & alpha & $[10^{-4},\,100]_{\log}$ \\
    \midrule
    Ridge\,Regression & alpha & $[10^{-4},\,100]_{\log}$ \\
    \bottomrule
  \end{tabular}
\end{table*}

Table~\ref{table:classical-ml-hyperparameters} lists the search spaces of all classical ML algorithms considered in this study. Sampling rules and the PCA or partial least squares component search (2--32) are identical to those used for the ANNs.


\appsection{Implementation details}
\label{appendix:implementation-details}

All ANN models employed in this study utilize reference implementations of their core architectures. For the model architecture implementations, we use the authors' reference code provided with the respective publications for ModernNCA \citep{ye2025revisiting}, TabR \citep{gorishniy2023tabr}, TabM \citep{gorishniy2025tabm}, FT-Transformer \citep{gorishniy2023revisiting}, T2G-Former \citep{yan2023t2gformer}, AMFormer \citep{cheng2024arithmetic}, and ExcelFormer \citep{chen2024excelformer}. For TabPFN \citep{hollmann2025accurate}, we use the authors' implementation including the pre-trained model weights. MLP follows the implementation from \citet{gorishniy2025tabm} and AutoInt follows the PyTorch reimplementation from \citet{gorishniy2023revisiting}. RealMLP was reimplemented based on the tabular benchmarking framework from \citet{holzmuller2025better}, excluding their custom training pipeline. Numerical feature embeddings were based on \texttt{rtdl-num-embeddings 0.0.12}, using an implementation of Quantile-Based Piecewise Linear Encoding that corresponds to the "Q-L" variation from Table 2 in \citet{gorishniy2022embeddings}. The benchmarking framework, including our unified model training interface, training loop with early stopping, data preprocessing pipelines (scaling, PCA, numerical embeddings), optimization utilities, cross-validation structure, YAML-based configuration system, and experimental logging, was implemented as part of this work to ensure consistent evaluation across all models.

The evaluation framework was developed in \texttt{Python 3.10} using \texttt{PyTorch 2.8.0} \citep{paszke2019pytorch} for deep learning models, \texttt{scikit-learn 1.6.1} \citep{pedregosa11a2011sklearn} for preprocessing and classical methods, \texttt{XGBoost 3.0.3} \citep{chen2016xgboost} for GBDT implementation, and \texttt{Optuna 4.4.0} \citep{akiba2019optuna} for hyperparameter optimization. The complete implementation, including hyperparameters and experimental results, is openly available (see Code and data availability in the main manuscript).


\appsection{Glossary}
\label{appendix:glossary}

This glossary provides brief and condensed, application-oriented notions of key Artificial Neural Network (ANN) terms as used in this study. For a detailed discussion, we refer to~\citep{bengio2017deep,aggarwal2018neural}.

\textbf{Activation function} -- a (typically) non-linear function applied within a layer that enables the model to (then) capture non-linear relationships.

\textbf{Architecture} -- the overall structural design of an  ANN, including which layers it contains, how they are connected (e.g., residual/skip connections), and how information flows from inputs to outputs.

\textbf{Attention} -- a mechanism that assigns data-dependent importance weights to a set of inputs and forms a weighted combination of them.

\textbf{Batch} -- a subset of the data processed together in a single forward and backward pass through the ANN.

\textbf{Embedding} -- a dense vector representation of an item in a continuous feature space.

\textbf{Epoch} -- one complete pass of all training data through the ANN.

\textbf{Feed-forward network} -- an ANN whose computation graph is a directed acyclic graph, with no cycles or feedback connections.

\textbf{Hyperparameters} -- variables that control the learning process and are chosen by the practitioner rather than learned during training.

\textbf{In-context learning} -- making predictions by conditioning on examples provided at inference time, without updating the ANN’s parameters.

\textbf{Layer} -- a building block of an ANN that applies a specific transformation to its input(s).

\textbf{Learning rate} -- a hyperparameter used by the optimizer that controls the step size of parameter updates during training.

\textbf{Loss function} -- the objective minimized during ANN training.

\textbf{Multi-head self-attention} -- multiple self-attention heads computed in parallel on learned projections of the inputs.

\textbf{Multilayer perceptron} -- a feed-forward, fully connected neural network composed of stacked linear (affine) layers with non-linear activations; often abbreviated as MLP.

\textbf{Optimizer} -- the algorithm that adjusts the ANN’s parameters to minimize the loss function during training.

\textbf{Parameters} -- internal variables (e.g., weights and biases) that are learned during training.

\textbf{Prior-data Fitted Network} -- a network trained on many synthetic datasets sampled from a prior distribution over supervised learning tasks. At inference time, it approximates Bayesian inference on new datasets in a single forward pass without parameter updates, enabling in-context learning.

\textbf{Representation learning} -- the process of automatically learning useful feature representations from raw inputs.

\textbf{Residual connection} -- a common form of skip connection that adds a block’s input to its output (element-wise), assuming matching dimensions.

\textbf{Self-attention} -- attention applied over elements within the same data sample.

\textbf{Skip connection} -- a connection that passes information from an earlier layer directly to a later one, bypassing intermediate layers.

\textbf{Tokenization} -- converting raw inputs into discrete units called tokens.

\textbf{Transformer} -- an ANN architecture built from stacks of multi-head self-attention and feed-forward blocks, connected with residual connections and normalization, usually with positional encodings to represent order.

\end{document}